\documentclass[10pt, a4paper]{article}
\usepackage{lrec}
\usepackage{graphicx}
\usepackage{tabularx}
\usepackage{soul}
\usepackage{url}
\usepackage{color}
\usepackage{longtable}

\usepackage{epstopdf}
\usepackage[latin1]{inputenc}

\usepackage{hyperref}
\usepackage{xstring}
\usepackage{lipsum}
\usepackage{amsmath}

\providecommand{\shortcite}[1]{\cite{#1}}

\newcommand{\indentm}{\hspace{2em}}

\title{ScienceExamCER: A High-Density Fine-Grained Science-Domain Corpus \\for Common Entity Recognition}

\name{Hannah Smith, Zeyu Zhang, John Culnan, Peter Jansen}

\address{School of Information, University of Arizona\\
         Tucson, Arizona, USA \\
         pajansen@email.arizona.edu\\
		}

\abstract{
Named entity recognition identifies common classes of entities in text, but these entity labels are generally sparse, limiting utility to downstream tasks.  In this work we present ScienceExamCER, a densely-labeled semantic classification corpus of 133k mentions in the science exam domain where nearly all (96\%) of content words have been annotated with one or more fine-grained semantic class labels including taxonomic groups, meronym groups, verb/action groups, properties and values, and synonyms.  Semantic class labels are drawn from a manually-constructed fine-grained typology of 601 classes generated through a data-driven analysis of 4,239 science exam questions.  We show an off-the-shelf BERT-based named entity recognition model modified for multi-label classification achieves an accuracy of 0.85 F1 on this task, suggesting strong utility for downstream tasks in science domain question answering requiring densely-labeled semantic classification. 
 \\ \newline  }

\begin{document}

\maketitleabstract

\section{Introduction}

%
%
\begin{figure*}[!t]
	\centering
	\includegraphics[scale=3.05]{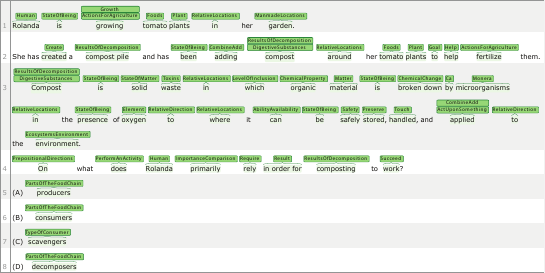}
	\caption{An example standardized science exam question densely annotated with one or more fine-grained semantic categories for nearly each word.  This 4-choice multiple choice question (here, under the curriculum topic \textit{``The Interdependence of Life \textgreater The Food Chain \textgreater Decomposers''}) is one of 4,239 drawn from the ARC corpus and densely annotated in this work. \label{fig:example}}
\end{figure*}

Named entity recognition (NER) \cite{grishman1996message} is a common natural language processing task that aims to abstract or categorize common classes of noun phrases in text, such as identifying ``Arthur'' as a \textit{person} or ``Montreal'' as a \textit{location}.  This high-level categorization of important entities in text is a staple of most modern NLP pipelines, 
and has a variety of applications for higher-level tasks including information extraction \cite{valenzuela2016odin}, knowledge base population \cite{dredze2010entity}, and question answering \cite{abujabal2017automated}.

\indentm Named entity recognition identifies common classes of entities in text, but these entity labels are generally sparse (typically occurring for between 10\% to 20\% of words in a corpus, see Section~\ref{sec:mentiondensity}), limiting utility to downstream tasks. In this work, we introduce the idea of common entity recognition (CER), which aims to tag all content words in text with an appropriate fine-grained semantic class. CER allows text to be automatically annotated with a much richer set of semantic labels, potentially providing greater utility for downstream applications.  We explore CER in the context of scientific text, and present ScienceExamCER, a training corpus annotated with over 113k common entity annotations drawn from a fine-grained set of over 600 semantic categories, which include named entities, as well as verb groups, properties and quantities, thematic types, and synonyms for key terminology.  We also release an off-the-shelf NER tagger modified to perform multilabel CER tagging. This BERT CER tagger achieves an accuracy of 0.85 F1 on this task, indicating that the tag ontology labels are well-defined and clearly identifiable. These two resources offer new opportunities to explore the impact of dense semantic annotation in downstream tasks.


\indentm We believe the notion of dense fine-grained semantic tagging to be potentially useful to any application domain, but explore common entity recognition here in the context of scientific text aimed at teaching and evaluating scientific knowledge. 
An example of this dense semantic classification in the context of standardized science exams is shown in Figure~\ref{fig:example}.  While CoreNLP \cite{manning2014stanford} does not locate any entities in the sentence \textit{``Rolanda is growing tomato plants in her garden''}, our CER annotation and system abstracts this sentence to \textit{``[Rolanda]$_\textit{Human}$ [is]$_\textit{StateOfBeing}$ [growing]$_\textit{Growth/ActionsForAgriculture}$ [tomato]$_\textit{Food}$ [plants]$_\textit{Plant}$ [in]$_\textit{RelativeLocation}$ her [garden]$_\textit{ManmadeLocation}$''}. 

\indentm We detail corpus and ontology/typology construction in Section 3, including a comparison of mention density with other common corpora.  Automated evaluations of CER performance are shown in Section 4, including an analyses of the training data requirements of this fine-grained classification, as well as an error analysis.

\section{Related Work}




\indentm Common sets of entity labels (or \textit{typologies}) have expanded from early experiments with a single label, \textit{organization} \cite{rau1991extracting}, to the 7 common MUC-6 types \cite{grishman1996message} typically used by NER systems, including named entities (\textit{person, organization, location}), temporal mentions (\textit{date, time}), and numeric categories (\textit{money, percent}).  Subsets of the MUC-6 types have been included in the typologies of benchmark NER corpora, including CoNLL-2003 \cite{sang2003introduction}, OntoNotes \cite{weischedel2013ontonotes}, and BBN \cite{weischedel2005bbn}.

\indentm Sekine et al. \shortcite{Sekine2002ExtendedNE} proposed an extended hierarchy of MUC-6 types expanded to include 150 open-domain category labels.  While most of these category labels are named entities, Sekine et al. include 10 measurement categories (e.g. \textit{weight, speed, temperature}) and 3 high-level natural object categories (\textit{animal, vegetable, mineral}) that most closely relate to the 601 fine-grained science categories in this work.  A subsequent version, the Extended Named Entity (ENE) Ontology \cite{sekine-2008-extended}, expands the typology to 200 classes, including 19 fine-grained expansions of the \textit{natural\_object} type, such as \textit{bird} or \textit{reptile}, as well as adding 5 meronym categories, such as \textit{plant\_part}, that further relax the working definition of named entities from proper names to include other categories \cite{nadeau2007survey}. 

\indentm While open-domain typologies are common, domain-specific typologies and corpora are also popular, occasionally making use of existing domain ontologies to reduce the burden in manually generating fine-grained typologies, such as the manual creation of the fine-grained science-domain typology in this work. An extreme example of fine-grained NER is the MedMentions corpus \cite{murty2018hierarchical}, which contains 246k mentions labelled with Universal Medical Language System (UMLS) \cite{bodenreider2004unified} categories, a fine-grained ontology of over 3.5 million medical concepts. 
Similarly, large knowledge bases can be filtered to automatically produce fine-grained typologies (as in FIGER \cite{ling2012fine} and HYENA \cite{yosef2012hyena}), or used to bootstrap the entity classification process in manually-generated typologies. Magnini et al. \shortcite{magnini-etal-2002-wordnet} demonstrate combining WordNet predicates \cite{fellbaum1998wordnet} with approximately 200 handcoded rules can achieve an F1 score of 0.85 on recognizing 10 common entity types, while  
Ritter et al. \shortcite{ritter2011named} use distantly supervised topic modeling over Freebase entities  \cite{bollacker2008freebase} to perform named entity recognition on social media posts, achieving an F1 score of 0.59 on 10 common entity types. 
With respect to larger typologies, Del Corro et al. \shortcite{del2015finet} perform super-fine grained entity typing using the 16k fine-grained WordNet types under the high-level taxonomic categories of \textit{person, organization, and location}, achieving a manually-evaluated precision of 59.9\% on the CoNLL corpus and 28.3\% on New York Times news articles.  For smaller manually-generated typologies, Mai et al.~\shortcite{mai2018empirical} demonstrate a model combining LSTMs, CNNs, CRFs, and dictionary-based methods can achieve an F1 of 83.1 on an in-house corpus labeled with Sekine's \shortcite{sekine-2008-extended} 200-class ENE ontology. 

\indentm NER has historically been approached using a wide variety of methods, including rules \cite{hanisch2005prominer}, feature-based machine learning systems \cite{Mayfield:2003:NER:1119176.1119205}, conditional random fields \cite{greenberg2018marginal}, contextualized embeddings \cite{peters2018deep}, and combinations thereof. 
Qu et al. \shortcite{qu-etal-2016-named} demonstrate that it is possible to use a conditional random field model to transfer NER performance between datasets, at least in part.  Ma et al. \shortcite{ma2016label} show embedding models can transfer performance in zero-shot settings on fine-grained named entity classification.  Expanding on this, recent transformer models \cite{peters2018deep,devlin2018bert} have shown strong transfer performance on a variety of text classification tasks including named entity recognition using large pretrained contextualized embeddings that are fine-tuned on comparatively small in-domain corpora.  In this work we make use of an off-the-shelf bidirectional transformer (BERT) NER system modified to support multi-label classification, and demonstrate strong performance on the fine-grained common entity recognition task. 




\section{Data and Annotation}

%
%
\begin{table}
\centering
\small
\begin{tabular}{lcc}
\textbf{Label}	 			&	\textbf{Examples}	&	\textbf{Prop.}	\\
\hline
StateOfBeing                &   is, are, be             &   4.6\%   \\
LevelOfInclusion            &   which, each, only       &   4.0\%   \\
RelativeLocation            &   inside, under           &   3.0\%   \\
Comparison                  &   identical, difference   &   1.9\%   \\
RelativeDirection           &   forward, upward         &   1.8\%   \\
ProbabilityAndCertainty     &   likely, possible        &   1.6\%   \\
Cause                       &   because, due to         &   1.5\%   \\
AmountComparison            &   most, more, less        &   1.5\%   \\
RelativeTime                &   during, after           &   1.4\%   \\
Creation                    &   produce, make, form     &   1.2\%   \\
PhasesOfWater               &   steam, ice             &   1.1\%   \\
CardinalNumber              &   one, 100                &   1.0\%   \\
ContainBeComposedOf         &   made of, contains       &   1.0\%   \\
IncreaseDecrease            &   increasing, decline     &   0.9\%   \\
Element                     &   oxygen, carbon          &   0.9\%   \\
Plant                       &   tree, crops, weeds      &   0.8\%   \\
Move                        &   placed, motion, travel  &   0.8\%   \\
Use                         &   with, apply             &   0.8\%   \\
AmountChangingActions       &   deplete, extend         &   0.8\%   \\
RelativeNumber              &   many, some, high        &   0.7\%   \\
Temperature                 &   hot, warm, cold         &   0.7\%   \\
Energy                      &   kinetic energy, power   &   0.7\%   \\
CombineAdd                  &   add, absorb, mix        &   0.7\%   \\
LiquidMatter                &   water, oil, droplets    &   0.7\%   \\
Scientist                   &   geologist, Galileo      &   0.6\%   \\
Quality                     &   best, good, useful      &   0.6\%   \\
Size                        &   large, thick, diameter  &   0.6\%   \\
AbilityAvailablity          &   potential, unable       &   0.6\%   \\
ManmadeObjects              &   ball, spoon, paper      &   0.6\%   \\
PrepDirections              &   on, through, along      &   0.6\%   \\
Human                       &   person, astronaut       &   0.6\%   \\
ActionsForAnimals           &   eat, migrate, swim      &   0.6\%   \\
InnerPlanets                &   earth, mars, venus      &   0.6\%   \\
QualityComparison           &   advantage, benefit      &   0.6\%   \\
Mammal                      &   dog, horse, bear        &   0.6\%   \\
Exemplar                    &   including, such as      &   0.5\%   \\
PlantPart                   &   leaves, flower, root    &   0.5\%   \\
PerformActivity             &   conduct, do             &   0.5\%   \\
Result                      &   effect, impact          &   0.5\%   \\
Compound                    &   carbon dioxide          &   0.5\%   \\
BodiesOfWater               &   ocean, lake, pond       &   0.5\%   \\
Help                        &   benefit, support        &   0.5\%   \\
Require                     &   need, must, takes       &   0.5\%   \\
Rock                        &   bedrock, boulder        &   0.5\%   \\
TemporalProperty            &   first, over time        &   0.5\%   \\
EarthPartsGross             &   surface, equator        &   0.5\%   \\
WeatherPhenomena            &   wind, cloud, drought    &   0.5\%   \\
Communicate                 &   explain, describe       &   0.5\%   \\
GeographicFormations        &   mountain, glacier       &   0.5\%   \\
ChangeInto                  &   become, converted       &   0.5\%   \\
Soil                        &   sand, ground, topsoil   &   0.4\%   \\
Color                       &   green, blue, white      &   0.4\%   \\
Star                        &   sun, proxima centauri   &   0.4\%   \\
PhaseChangingActions        &   melt, evaporation       &   0.4\%   \\
Nutrition                   &   food, nutrient          &   0.4\%   \\
AnimalPart                  &   body, organ, eye        &   0.4\%   \\
TimeUnit                    &   day, second, year       &   0.4\%   \\
Method                      &   process, procedure      &   0.4\%   \\
StateOfMatter               &   solid, liquid, gas      &   0.4\%   \\
\end{tabular}
\vspace{-4pt}
\caption{The most frequent subset of the 601 semantic classification labels used to annotate the ScienceExamCER corpus.  Proportion refers to the proportion of mentions in the training set that are labeled with a given category.  The full set of 601 classes is included in the supplementary material. \label{tab:examplesemanticcategories}}
\end{table}


\subsection{Corpus}
We annotate fine-grained semantic classes on standardized science exam questions drawn from the Aristo Reasoning Challenge (ARC) corpus \cite{clark2018think}, which contains 7,787 elementary and middle school ($3^{rd}$ through $9^{th}$ grade) standardized exam questions drawn from 12 US states.  Each question is a 4-way multiple choice question, ranging from short direct questions to detailed multi-step problems grounded in examples.  An example question is shown in Figure~\ref{fig:example}. Question text contains an average of 21 words across 1.7 sentences, while answer candidate text averages 4 words, but can be as short as a single word (as in Figure~\ref{fig:example}).   In this work we draw 4,239 questions from the ARC corpus, consisting of the full training and development folds, to use for our semantic labeling and prediction tasks. 

\subsection{Semantic Class Labels}

We conducted a large data-driven analysis of the 4,239 science exam questions with the aim of identifying a set of high-level semantic categories that would provide near total coverage for classifying or grouping nearly all of the 156k words found across the question and answer text in this corpus. 
While named entity recognition typically focuses on proper names with specific referents \cite{nadeau2007survey}, in the end we arrived at creating 601 fine-grained categories spanning 6 classes of groups: 

{\flushleft\textbf{Taxonomic Groups:}} High-level categories expressing taxonomic membership, such as that a \textit{hummingbird} is a kind of \textit{bird}.  This (or stricter interpretations) is the common form of entity classification in most named entity recognition corpora. 

{\flushleft\textbf{Meronym Groups:}} Categories expressing part-of relations, such as that a \textit{fin} is a part of an \textit{aquatic animal}, an \textit{x-axis} is a part of a \textit{graphical representation}, or that an \textit{individual} is a part of a \textit{group}. 

{\flushleft\textbf{Action Groups:}} Collections of action words that tend to describe similar ideas.  For example, \textit{decrease, increase, contract, expand, inflate, deflate, accelerate, decelerate, lower, raise} all describe a group of \textit{actions that involve increasing or decreasing quantities}. 

{\flushleft\textbf{Thematic word groups:}} Groups of words that surround a particular topic.  For example, \textit{observe, conduct an experiment, compare, study, consider, test, collect, record, gather, examine,} and \textit{research} are some of the words included in the \textit{performing research using the scientific method} semantic class.

{\flushleft\textbf{Properties and Values:}} Common science-domain properties of objects, such as \textit{mass, size,} or \textit{conductivity}, typically grouped with common values they might take, such as \textit{soft, brittle, or hard} in the case of \textit{hardness}. 

{\flushleft\textbf{Synonyms:}} Groups of words that tend to express similar ideas in the context of science exams.  For example, \textit{disease, infection, and sick} all convey the notion of \textit{illness}. 
\\

\indentm To identify specific instances of these categories in the science exam domain, we first sorted questions into fine-grained curriculum topics using the 406 detailed science-domain question classification labels of Xu et al. \shortcite{Xu2019MulticlassHQ}, noting that common categories of words tended to emerge upon detailed manual inspection when questions on similar topics were examined together.  We proceeded through several iterations of this process, recording candidate high-level semantic classes, as well as seed words that belonged to those categories.  After assembling a large list of candidate categories, we further enumerated the seed words with encyclopedic knowledge manually through web searches.  For example, while the annotators may have only observed the words \textit{Sun} and \textit{Proxima Centauri} in the corpus for the \textit{Star} category, we would manually expand this to also include other nearby stars such as \textit{Vega}, \textit{Polaris}, and \textit{Wolf 359}.  

\indentm As a final step, we automatically expanded the seed word list to include lexical variations of each manually added word by first using pretrained GLoVe embeddings \cite{pennington2014glove} to compute the top-N most similar words to a given seed word using cosine similarity, then using several low-precision high-recall heuristics to identify words that had the potential to be lexical variations of an existing word on the seed list.  We then generated a frequency histogram of any word present in the corpus that did not yet belong to at least one semantic category, and either placed it in an existing category, or formed a new category for that word and repeated the expansion process for seed words.  This detailed manual category development process required approximately three weeks of annotator time, ultimately arriving at a list of 601 high-level semantic categories, with an extensive list of both manually and automatically populated seed words for each category.  The full list of semantic categories and seed words is included in the \textit{supplementary material}. 


%
%
\begin{table*}
\centering
\begin{tabular}{lcccccc}
\textbf{Measure}	 						&	\textbf{ScienceExamCER}	&	\textbf{OntoNotes 5}&   \textbf{BBN}    	&   \textbf{GUM}    &	\textbf{CoNLL 2003}	\\
\hline
\textit{Entity Categories}                  &   601                     &   18                  &   64                  &   11              &   4                    \\
\textit{Total Mentions}                     &	~133k${^{*}}$	        &	162k			   	&   172k                &   11k             &	~35k					\vspace{1.5mm}\\

\textit{Words}								&	156k				    &	2.44M				&   1.05M               &   55k             &	264k				\\
~~\textit{Labeled Words}					&	117k				    &	284k				&   257k                &   33k             &	51k					\\
~~\textit{Mention Density (overall)}		&	\textbf{75\%}   	    &	12\%				&   25\%                &   59\%            &	~19\%				\vspace{1.5mm}\\

\textit{Content Words}						&	104k				    &	1.39M				&  677k                 &   34k             &	190k				\\
~~\textit{Labelled Content Words}			&	100k				    &	255k				&  243k                 &   21k             &	50k					\\
~~\textit{Mention Density (Content Words)}	&	\textbf{96\%}		    &	18\%				&  36\%                 &   62\%            &	~27\%				\vspace{1.5mm}\\


\end{tabular}
\caption{Summary statistics including \textit{mention density} for the ScienceExamCER corpus, as well as four other common benchmark corpora.  At 96\%, the ScienceExamCER is significantly more  densely labeled than the next-nearest corpus. (\textit{*} denotes that approximately 16k spans have multiple labels, and as such the total mentions exceeds the total labeled words). \label{tab:summarystatistics}}
\end{table*}

%
%
\begin{table}[]
    \centering
    \begin{tabular}{lcc}
         \textbf{Fold}  &   \textbf{Science Questions}  &   \textbf{Words}  \\
         \hline
         Train          &   2,696                &   108,396              \\
         Development    &   674                  &   27,560              \\
         Test           &   869                  &   35,379              \\
    \end{tabular}
    \caption{Summary statistics for the training, evaluation, and test sets used for evaluating semantic category classification. }
    \label{tab:summarystatisticsquestionset}
\end{table}

\subsection{Annotation Procedure}

Annotating a large set of semantic classes onto more than one hundred thousand words presents challenges with annotation consistency and tractability.  It would be challenging for crowdworkers to learn a detailed set of 601 fine-grained semantic categories, and extremely time consuming for research assistants to traditionally annotate a collection at this scale.  To overcome these challenges, we modified the annotation task to automatically preannotate the entire corpus using the large set of bootstrapped seed words associated with each semantic class, effectively preannotating each word with a set of possible semantic category labels.  These preannotated labels are effectively low-precision and high-recall, most often containing the correct label(s) for a given mention, but also containing other incorrect labels that must be manually removed by an annotator.  A total of 226k preannotated mentions were generated (an average of 1.5 per word), which was reduced to 133k mentions (0.9 per word) after incorrect labels were removed by the annotator.  We used the BRAT annotation tool \cite{stenetorp2012brat} for the label removal step.  To ease the annotator's need for switching semantic contexts, questions were presented to the annotator sorted by curriculum topic using the question classification annotation of Xu et al. \shortcite{Xu2019MulticlassHQ}.  The annotation procedure took approximately 2.5 minutes per question, for a total of 200 hours.

\indentm A clear question with this ``preannotate-then-filter'' annotation protocol is how well this procedure is able to provide both coverage and accurate labels for the words in the corpus.  Our analysis in Section~\ref{sec:mentiondensity} shows that after annotation, 96\% of content words and 75\% of all words have at least one gold semantic category label, suggesting this protocol allows for near-complete coverage of content words at a fraction of the time required to make accurate 601-class annotation judgements at scale.  Both our interannotator agreement (included below) and automatic classification performance are high, suggesting adequate annotated label accuracy.


{\flushleft\textbf{Label distribution:}} Named entity corpora often have many labels in their typologies, but the majority of mentions tend to cluster around a small set of possible labels \cite{choi-etal-2018-ultra}.  The distribution of most frequent labels after annotation is shown in Table~\ref{tab:examplesemanticcategories}. The usage of the 601 total semantic class labels in this corpus is well distributed, with the 356 most-frequent types covering 95\% of the total mentions, while 479 types cover 99\% of mentions.  At the 99\% level, categories (for example, \textit{Geometric Qualities}, such as \textit{angle, slope, or circumference}) still contain 16 mentions, highlighting the scale of the corpus.


{\flushleft\textbf{Interannotator agreement:}} Each question was annotated by a single annotator.  A second annotator was trained in the annotation procedure and re-annotated 50 questions totalling 1,756 tokens.  Between both annotators, a total of 1,369 mentions were annotated with semantic class labels.  Total percent agreement across both annotators was 76\%.\footnote{Because the bootstrapped preannotation procedure reduces the set of possible labels for a given mention from 601 to an average of approximately 2 (the average number of preannotated labels per annotated word), Cohen's Kappa \cite{cohen1960coefficient} would either be artificially inflated (if treating the annotation as a 601 class labeling problem) or reduced (if treating annotation as a 2 class problem).  As such we report raw percent agreement, which (as critiqued by Cohen) has known problems when dealing with highly skewed frequency distributions of labels, particularly when few labels are present.  Here, the number of label categories is high, and (as shown in Table~\ref{tab:examplesemanticcategories}) the frequency of labels is well distributed across the label set.  As such, the inflation of the percent agreement statistic is likely to be minimal. }  Upon inspection, labeling multi-word sequences as either a single mention or multiple smaller mentions was a frequent source of disagreement.  When these cases were removed, percent agreement rose to 83\%.

~\\



\subsection{Mention density comparison}
\label{sec:mentiondensity}

To increase the utility of our common entity corpus for downstream tasks, one of the design goals was to provide at least one high-level semantic category to nearly every word in the corpus.  To measure this we define the notion of the \textit{mention density} of a corpus as the proportion of words that contain at least one entity label.\footnote{Specifically, the proportion of non-punctuation tokens in a BIO-formatted corpus that are labelled with either a B (beginning) or I (inside) tag.} 
We compare the mention density of this corpus with the English subsets of the four benchmark named entity recognition corpora listed below:

{\flushleft\textbf{CoNLL}} \cite{sang2003introduction}: The CoNLL 2003 Named Entity Recognition Shared Task corpus, which includes 4 entity labels that are a subset of the MUC-6 typology: \textit{person, location, organization} and \textit{miscellaneous}. 
{\flushleft\textbf{OntoNotes 5.0}} \cite{weischedel2013ontonotes}: A large multi-genre corpus of news media, blog, newsgroup, and conversational text, annotated with 18 entity labels, including the MUC-6 types.
{\flushleft\textbf{BBN}} \cite{weischedel2005bbn}: A corpus of news text annotated with 21 course entity types, including 12 named entity types (e.g. \textit{person, organization, product}) and 7 numeric types (e.g. \textit{date, percent, cardinal number}).  The full set of entity labels includes 64 fine-grained types. 
{\flushleft\textbf{GUM}} \cite{Zeldes2017}: An open-domain corpus annotated with a collapsed set of OntoNote entities reduced to 11 entity types, such as \textit{person, organization,} or \textit{place}.  Two additional catch-all tags are added, \textit{object} and \textit{abstract}, which provide high-level but minimally informative categorical information for large noun phrases.  Approximately 40.5\% of the labelled words in this corpus are labelled as either \textit{object} or \textit{abstract}.\\

\indentm The analysis of mention density is shown in Table~\ref{tab:summarystatistics}. Overall, the mention density of this science corpus is 75\%, meaning that 75\% of all words in the corpus are annotated with at least one high-level semantic category.  When considering only content words (here, determined to be \textit{nouns, verbs, adjectives, adverbs, and numbers}), this proportion increases to 96\%.  The mention density for the named entity corpora examined in Table~\ref{tab:summarystatistics} ranges between 12\% and 59\% for all words, and 18\% to 62\% when considering only content words.  At 62\%, the GUM corpus contains the next-nearest mention density to the ScienceExamCER corpus, however a large portion of those mentions (40.5\%) of words) use the high-level \textit{object} or \textit{abstract} labels, and as such are of limited informativeness to downstream tasks.  BBN, the corpus with the next-nearest mention density to GUM, has labels for only 36\% of it content words, and 25\% of all words. 

\section{Experimental Results}

\subsection{Model}

Our semantic class labeling task is conceptually similar to named entity recognition or entity typing, only requiring a label for nearly every word in an input sentence.  In light of this, here we use an off-the-shelf named entity recognition model, and show it also performs well on the densely-labeled common entity recognition task. 

\indentm Recently, pretrained bidirectional encoder representation from transformer (BERT) models \cite{devlin2018bert} have shown state-of-the-art performance at both named entity recognition as well as a variety of other token-level classification tasks.  In this work, we use an off-the-shelf implementation of a BERT-based named entity recognition system, BERT-NER\footnote{\url{https://github.com/kamalkraj/BERT-NER}}.  Most approaches to named entity recognition model the task as a single-label prediction task, where each word has at most one label.  We modify the BERT-NER implementation to allow for multi-label predictions using the following method. 



\indentm Given a sentence $S$ consisting of $L$ tokens, such that $S=(x_1, x_2, ..., x_L)$, the original BERT-based token classification model generates $L$ respective $M$-dimensional encodings $(x_1, x_2, ..., x_L)$, one for each token. These encodings then pass through a \textit{softmax} layer and make use of a multi-classes cross entropy loss function that generates a single class prediction per token.  
We adapt this system to multi-label classification by using a \textit{sigmoid} function and binary cross entropy in place of the original loss function to allow the classifications for each token to return non-zero values for more than one class.  More formally, our loss function becomes:
\begin{equation}
\begin{aligned}
    L_{multilabel} = {}&-\frac{1}{M}\sum_{m=1}^{M}[\tilde{y}_l^m\cdot{log\sigma(x_l^m)} + {} \\
    &(1-\tilde{y}_l^m)\cdot{log(1-\sigma(x_l^m))}] \\
\end{aligned}
\end{equation}

\begin{equation}
\begin{aligned}
    \sigma(x_l^m)&=\frac{1}{1+e^{-x_l^m}}
\end{aligned}
\end{equation}
where $M$ is the number of total classes, $x_l$ is $M$-dimensional encoding for the $l$-th token in sentence, $\tilde{y}_l$ is the $l$-th token's gold label vector, and $\sigma$ is the \textit{sigmoid} activation function.




{\flushleft\textbf{Folds:}} Because of the expense associated with annotating a large corpus, only the training and development subsets of the ARC corpus were manually annotated with semantic class labels.  As such we repurpose the original development set for testing, and hold out 20\% of the training corpus for development. Summary statistics on these folds are provided in Table~\ref{tab:summarystatisticsquestionset}. 

{\flushleft\textbf{Hyperparameters:}} We make use of the pre-trained English \texttt{BERT-Base-cased} model\footnote{\url{https://github.com/google-research/bert}}, with a maximum sequence length of 64.  The threshold for the sigmoid activation layer was tuned on the development set, with a value of 0.4 found to provide good performance. The large number of possible class labels in our task compared with typical named entity recognition datasets, combined with the modified multi-label loss function, necessitated significantly longer training times for the model to converge.  We empirically found that the model tended to converge by 140 epochs, which took approximately 5 hours to train using dual RTX2080Ti GPUs.  Classification of the entire test dataset is comparatively fast, providing semantic class labels at a rate of approximately 900 questions (35,000 words) per minute, enabling the pre-trained model to be run on other science-domain corpora (for example, textbooks, study guides, Simple Wikipedia, or other grade-appropriate knowledge resources) at scale.  

\subsection{Evaluation}

%
%
\begin{table}
\centering
\begin{tabular}{lcccc}
\textbf{Model} 								& \textbf{Fold} 	&	\textbf{Prec.}  &   \textbf{Recall}     &   \textbf{F1} 	\\
\hline
BERT-NER                                    &   dev             &   0.84            &   0.85                &   0.84            \\
BERT-NER                                    &   test            &   0.85            &   0.86                &   0.85            \\

\end{tabular}
\caption{Performance on the 601-category fine-grained semantic classification task on the development and test folds using the BERT-NER model.  \label{tab:performance}}
\end{table}

The results for our semantic classification task on the ScienceExamCER corpus using the 601-class fine-grained typology are shown in Table~\ref{tab:performance}.  We evaluate entity classification performance using the standard definitions of Precision, Recall, and F1.  Overall classification performance is high, reaching 0.85 F1 on the held-out test set. This suggests the common entity recognition performance is sufficiently high to be useful for a variety of downstream tasks.  To further characterize performance, we investigate how the availability of training data affects this fine-grained classification task, as well as common classes of prediction errors the BERT-NER model makes. 





\subsection{Performance vs Training Data}

%
%
\begin{figure}[!t]
	\centering
	\includegraphics[scale=0.30]{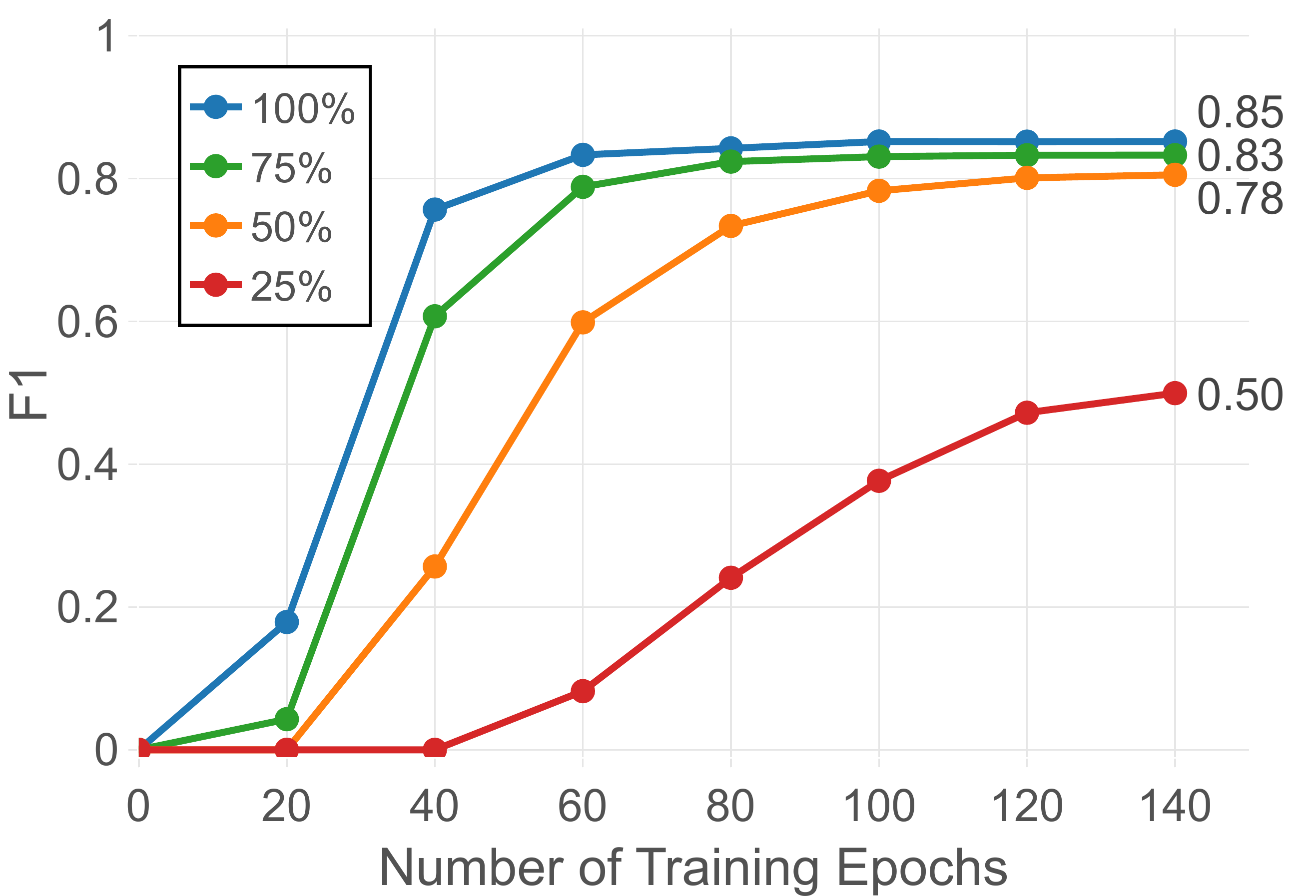}
	\caption{Classification performance (F1) versus the number of training epochs when training the model with less data.  Series represent training the model with the entire training set, or randomly subsampled proportions of training data summing to 75\%, 50\%, and 25\% of the original training set size.  Each point represents the average of 5 randomly subsampled training sets.  \label{fig:datadependence}}
\end{figure}

Manually annotating fine-grained mentions in large corpora is expensive and time consuming.  To investigate how classification performance varies with availability of training data, we randomly subsampled smaller training sets from our full training corpus that were 25\%, 50\%, or 75\% as large, corresponding to spending approximately 50, 100, or 150 hours at the manual annotation task, respectively.  The results are shown in Figure~\ref{fig:datadependence}.  With only 25\% of training data available, F1 performance decreases dramatically from 0.85 to 0.50. 50\% of training data decreases classification performance by 7 points, while 75\% of available training data decreases classification performance by 2 points.  This suggests that the scale of training data generated in this work is provides near saturated performance using the BERT-NER model, and that annotating the remainder of available standardized science exam questions in the ARC corpus would likely result in only a minimal increase on classification performance. 


\subsection{Error Analysis}

%
%
\begin{table}
\centering
\begin{tabular}{lc}
\textbf{Error Class} 									&	\textbf{Prop.}	\\
\hline
\textit{Predicted label also good}						&	24\%	\\
\textit{Model did not generate prediction}				&	24\%	\\
\textit{Multiple gold labels, one found}				&	21\%	\\
\textit{Predicted label semantically near gold label}	&	17\%	\\
\textit{Gold label incorrect}							&	7\%		\\
\textit{Multi-word Expression}							&	6\%		\\
\textit{Predicted label using incorrect word sense}		&	5\%		\\

\end{tabular}
\caption{An analysis of common categories of model prediction errors, as a proportion of the first 100 errors on the test set.  Note that a given errorful prediction may belong to more than one category, and as such the proportions do not sum to 100\%. \label{tab:erroranalysis}}
\end{table}

To better understand the sources of error in our model, we conducted an analysis of the first 100 errorful predictions on the test set, with the results shown in Table~\ref{tab:erroranalysis}.  Nearly one third of errors are due to issues with the annotation, such as a mention missing an additional label that is also good (24\% of errors), or the manually annotated gold label being incorrect (7\% of errors).  For a substantial portion of errors (24\%), no single semantic class rose to meet the activation threshold of the sigmoid layer and the model did not produce a prediction for that word, while, similarly, in 21\% of cases only one label of a multi-label word was produced.  The remaining errors broadly cluster around technical challenges in determining the semantics of each category, including word-sense disambiguation (5\% of errors), locating multi-word expressions (6\% of errors), or predicting a label whose category is semantically similar to the gold label (17\% of errors).

\section{Conclusion}

We present ScienceExamCER, a densely annotated corpus of science exam questions for common entity recognition where nearly every word is annotated with fine-grained semantic classification labels drawn from a manually-constructed typology of 601 semantic classes.  We demonstrate that BERT-NER, an off-the-shelf named entity recognition model, achieves 0.85 F1 on classifying these fine-grained semantic classes on unseen text in a multi-label setting.  The data and code are released with the goal of supporting downstream tasks in question answering that are able to make use of this dense semantic category annotation.  

\section{Supplementary Material}

The annotated corpora, fine-grained typology, and pretrained models for this work are available at \url{http://cognitiveai.org/explanationbank/}.  A truncated version of the typology is included in the \texttt{Appendix} below.

\section{Acknowledgements}
This work was supported by the Allen Institute for Artificial Intelligence and the National Science Foundation (NSF Award \#1815948, ``Explainable Natural Language Inference'', to PJ).  We thank Peter Clark for thoughtful discussions on this work and comments on an earlier draft. 

\section{Bibliographical References}
\label{main:ref}

\bibliographystyle{lrec}
\bibliography{refs}


\section*{Appendix}

The full list of semantic category labels is included in Table~\ref{tab:semanticcategoriesfulllist} below. 

\clearpage
\onecolumn
%
%
\centering
\small
\begin{longtable}{ll}
\textbf{Label}	 			&	\textbf{Examples}	\\
\hline
Celestial Object	&	celestial object, astronomical body, celestial body, extrasolar-body	\\
~~     Asteroid	&	asteroid belt, Kuiper belt, asteroid, planetoid, Iris, Flora, Metis, Hygiea	\\
~~  Black Hole	&	super-massive black hole, black hole, Centaurus A, Sagittarius A*	\\
~~  Comet	&	Halley's comet, comet, Shoemaker-Levy, Great Comet of 1807	\\
~~  Constellation	&	constellation, Leo, Little Dipper, pattern of stars, star pattern, ursa Major	\\
~~  Galaxy	&	Andromeda galaxy, Milky Way, M87, galaxy, Large Magellanic Cloud	\\
~~~~ Galaxy Parts	&	galactic region, halo, spiral arm, spiral arms, nuclear bulge	\\
\end{longtable}

\end{document}